# Front-propagation Algorithm: Explainable AI Technique for Extracting Linear Function Approximations from Neural Networks

Javier Viaña[1][0000-0002-0563-784X]

[1] MIT Kavli Institute for Astrophysics and Space Research,
Massachusetts Institute of Technology, Cambridge 02139, USA
`vianajr@mit.edu`

**Abstract.** This paper introduces the front-propagation algorithm, a novel eXplainable AI (XAI) technique designed to elucidate the decision-making logic of deep neural networks. Unlike other popular explainability algorithms such as Integrated Gradients or Shapley Values, the proposed algorithm is able to extract an accurate and consistent linear function explanation of the network in a single forward pass of the trained model. This nuance sets apart the time complexity of the front-propagation as it could be running real-time and in parallel with deployed models. We packaged this algorithm in a software called `front-prop` and we demonstrate its efficacy in providing accurate linear functions with three different neural network architectures trained on publicly available benchmark datasets.

**Keywords:** Explainable AI (XAI), Deep Learning, Neural Networks, Integrated Gradients, Shapley Values, Model transparency.

## 1   Introduction

Neural networks have demonstrated remarkable performance across a myriad of tasks, ranging from image recognition to natural language processing [1]. However, the inherent complexity of these models often renders their decision-making processes opaque and difficult to interpret, hindering their deployment in critical domains where transparency and trustworthiness are paramount. In recent years, the pursuit of eXplainable AI (XAI) techniques has emerged as a crucial endeavor to demystify the inner workings of such black-box models [2].

Among the plethora of XAI methodologies, Integrated Gradients has garnered significant attention for its effectiveness in attributing feature importance to model predictions [3]. By systematically perturbing input features and integrating the gradients of model outputs with respect to these perturbations, Integrated Gradients provides insightful explanations for individual predictions, thereby enhancing model interpretability.

In this paper, we delve into the realm of explainable AI tools, particularly focusing on a technique akin to Integrated Gradients, to facilitate the extraction of linear function approximations from neural networks. The rationale behind this endeavor lies in the interpretability and simplicity afforded by linear models [4], which stand in stark contrast to the complexity of neural networks. By distilling the essence of neural network behaviors into linear approximations, we aim to bridge the gap between model complexity and interpretability.

This paper contributes to the burgeoning field of XAI by presenting a novel methodology tailored to extract linear function approximations from neural networks, thereby shedding light on their decision-making processes. We showcase the efficacy of our approach through empirical evaluations on benchmark datasets, demonstrating the utility of linear approximations in enhancing model interpretability without sacrificing performance.

In addition to Integrated Gradients there are currently several other ways to provide explanations for the predictions of complex machine learning models. To list some:

- Saliency Maps [5], which highlight the most relevant regions of the input data that contribute to the model's prediction. They are computed by taking the gradient of the output with respect to the input, effectively identifying the features that have the greatest impact on the model's decision.
- Gradient-weighted Class Activation Mapping (Grad-CAM) [6], which generates class-discriminative visual explanations by computing the gradients of the target class score with respect to the feature maps of a convolutional neural network. This technique highlights the regions in the input image that are most relevant to the predicted class.
- Layer-wise Relevance Propagation (LRP) [7], which decomposes the prediction of a neural network by attributing relevance scores to each neuron in the network's hidden layers. These relevance scores indicate the contribution of each neuron to the final prediction, thereby providing insights into the model's decision-making process.
- Shapley Values [8], an idea that originates from cooperative game theory and aim to fairly distribute the contribution of each feature to the model's output. In the context of explainable AI, Shapley values quantify the marginal contribution of each feature to the prediction, providing a globally consistent explanation for individual predictions.
- SmoothGrad [9], which improves the interpretability of gradient-based attribution methods by reducing noise in the gradient signal. It computes the average gradient across multiple noisy samples of the input data, resulting in smoother and more reliable explanations for the model's predictions.
- Local Interpretable Model-agnostic Explanations (LIME) [10], which generates local explanations for individual predictions by fitting an interpretable model (e.g., linear regression) to perturbed instances of the input data. By approximating the model's behavior in the vicinity of a particular prediction, LIME provides insights into the decision-making process of complex machine learning models.

These techniques form a diverse toolbox of explainable AI methods that facilitate the interpretation of complex machine learning models. Which depending on the specific characteristics of the model and the interpretability requirements of the

## 2 Methodology

The proposed front-propagation algorithm aims to extract a linear function approximation of the behavior of a neural network in the vicinity of a given instance, denoted as the base instance. As usual with other distillation techniques, the goal of this algorithm is to generate a simple function that can replace the model while still performing well for other datapoints that are "near" the base instance. The key difference with other methods, however, is the low computational cost that is required by the proposed front-propagation algorithm to obtain such function.

This paper explains the theoretical framework that underlies the aforementioned algorithm and discusses its potential benefits for the AI community. Indeed, the information of the linear function approximation is very relevant in various applications, not only to quantify the contributions of each of the input dimensions, but also to determine if the network is following a biased or an outlier reasoning, to supervise the model's behavior, or simply to extract relevant knowledge of the problem.

To describe the inner workings of the algorithm, we first consider a simple feed-forward neural network with $H$ hidden layers that performs a transformation from a $N_{in}$-dimensional input space $\mathbf{x} = \{x_1, x_2, \ldots, x_{N_{in}}\}$ into a $N_{out}$-dimensional output space $\hat{\mathbf{y}} = \{\hat{y}_1, \hat{y}_1, \ldots, \hat{y}_{N_{out}}\}$, as depicted in Fig.1. If $N_{out}$ was 1, then we would only need to extract a single linear function approximation of the network. However, in a generic network, we can expect to look for $N_{out}$ linear functions, each of them associated to each of the output dimensions. We start from a specific base input instance $\mathbf{x_i}$, and the associated output of the network $\hat{\mathbf{y}}_i$. In order to obtain the coefficients of these linear approximations, we will traverse all the layers sequentially in a forward-pass from the input layer to the output layer. Indeed, at every layer we will be calculating the linear function approximations that replace the network till that point. In other words, if a given hidden layer, which we denote by $l_h$, has $N_h$ neurons, we should be able to obtain $N_h$ linear functions. The information of these linear functions at $l_h$ can simply be stored in a matrix of dimension $N_h \times N_{in}$, which we denote by $\mathbf{M_h}$, where the entries of the matrix represent the coefficients that multiply each of the input dimensions. Therefore, the goal at every layer is to get these matrix coefficients. However, to do so we first need to compute the gradient of the layer's outputs with respect to the layer's inputs. If we have a fully connected dense layer, this is simply the derivative of the layer's activation functions evaluated at the base instance.

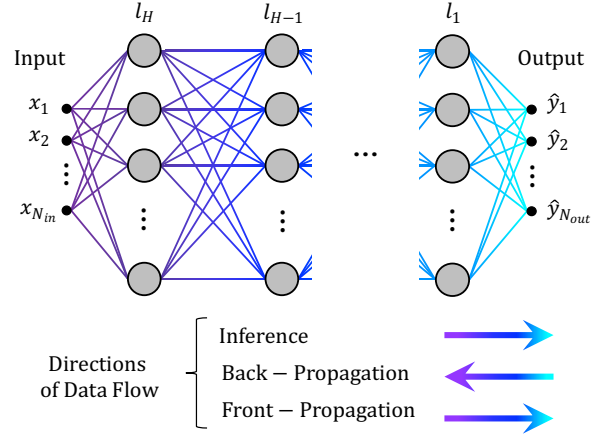

**Fig. 1.** Generic feed-forward neural network architecture. Difference in the flow of data in inference, back-propagation, and front-propagation.

At this point, one can start seeing the similarities between the proposed algorithm and the back-propagation algorithm:

- The back-propagation algorithm calculates the gradient of the loss function with respect to the weights and biases, and then corrects those parameters to optimize the network. In this algorithm the information flows from the output layer towards the input layer.
- The proposed front-propagation algorithm on the other hand calculates the gradient of the outputs with respect to the inputs, which then serve as an explanation of the network's reasoning. In this algorithm the information flows from the input layer towards the output layer.

Both the back and front-propagation algorithms are triggered after the network has made an inference. In the first case to correct the network's parameters and in the second to understand the reasoning of the network. Both algorithms share the same time complexity, and only require to traverse once the network to obtain the desired results, which makes the front-propagation very attractive for fast computations.

The name front-propagation was inspired on the similarities with the back-propagation algorithm and the fact that this algorithm traverses the network in the opposite direction.

## 3 Algorithm

Let us consider a generic $\gamma$-th neuron of an arbitrary intermediate layer $l_h$, as shown in Fig. 2. The value passed to this neuron is the sum of the product between the output vector of layer $l_{h+1}$, denoted $\mathbf{t}_{(h+1)}$, and the corresponding weight vector $\mathbf{w}_{h,\gamma}$, plus the bias $b_{h,\gamma}$. The output of the studied neuron, denoted by $t_{(h)_\gamma}$, is then obtained by mapping this value with the activation function $g_{h,\gamma}$. In formulation, $t_{(h)_\gamma} = g_{h,\gamma}\left(b_{h,\gamma} + \sum_{j=1}^{N_h} w_{h,\gamma,j} \cdot t_{(h+1)_j}\right)$. This value $t_{(h)_\gamma}$ is the $\gamma$-th entry of the vector $\mathbf{t_{(h)}}$, which denotes the output vector of layer $l_h$.

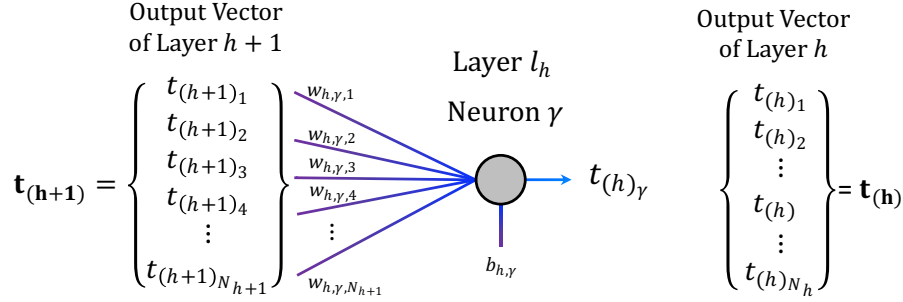

**Fig. 2.** Transformation of data carried out inside a generic neuron of the network.

We start with the neurons of the network's first hidden layer, $l_H$. For a given $\alpha$-th neuron of this layer $l_H$, the output $t_{(H)_\alpha}$ can be expressed as $g_{H,\alpha}\left(b_{H,\alpha} + \sum_{j=1}^{N_{in}} w_{H,\alpha,j} \cdot x_j\right)$. Instead of substituting $x_j$ with the corresponding values of the instance studied, we preserve $x_j$ as a placeholder variable that identifies the $j$-th input dimension. We remind the reader that the goal is to find the linear dependencies that relate the input and the output spaces, therefore, we do not want to substitute the values of $\mathbf{x}$.

We denote the argument that is passed to the activation function by $s(\mathbf{x})$, which for the case of the neuron studied is $s_{H,\alpha}(\mathbf{x}) = b_{H,\alpha} + \sum_{j=1}^{N_{in}} w_{H,\alpha,j} \cdot x_j$. At this point, the term $s_{H,\alpha}(\mathbf{x})$ is in fact a linear combination of the input dimensions, and the activation function is the one responsible for introducing the non-linearity in the obtention of $t_{(H)_\alpha}$. In order to find the linear dependencies between $t_{(H)_\alpha}$ and the input dimensions $x_j$, we first need to get the derivative of the activation function at the base instance. This is of course the tangent line that approximates the activation function at the base instance. We denoted our base instance by $\mathbf{x_i}$, thus, the representation of this point instance in the dimensions of the activation function is identified by the coordinates $\{s_{H,\alpha}(\mathbf{x_i}), t_{(H)_\alpha}(\mathbf{x_i})\}$, as shown in Fig. 3.

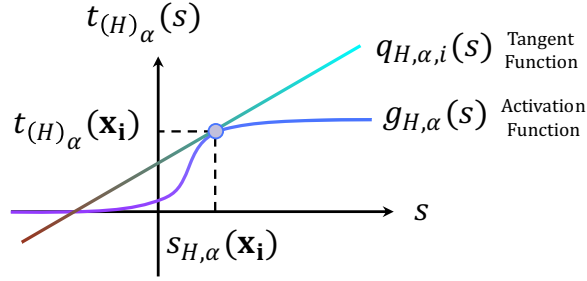

**Fig. 3.** Linear function approximation of the activation function around the base instance.

The tangent line of the activation function $g_{H,\alpha}$ at the base instance can be expressed as

$$q_{H,\alpha,i}(s) = m_{H,\alpha,i} \cdot s + n_{H,\alpha,i}. \tag{1}$$

where $q_{H,\alpha,i}\left(s_{H,\alpha}(\mathbf{x_i})\right) = g_{H,\alpha}\left(s_{H,\alpha}(\mathbf{x_i})\right)$ and $q_{H,\alpha,i}\left(s_{H,\alpha}(\mathbf{x_i})\right) = m_{H,\alpha,i} \cdot s_{H,\alpha}(\mathbf{x_i}) + n_{H,\alpha,i}$. The calculation of the slope of this linear function is $m_{H,\alpha,i}$ and is calculated simply by applying the derivative of the activation function and evaluating it at the base instance, i.e.,

$$m_{H,\alpha,i} = \left.\frac{dg_{H,\alpha}(s)}{ds}\right|_{s_{H,\alpha}(\mathbf{x_i})}. \tag{2}$$

The calculation of the independent term of the linear function, $n_{H,\alpha,i}$, is the obtained by

$$n_{H,\alpha,i} = q_{H,\alpha,i}\left(s_{H,\alpha}(\mathbf{x_i})\right) - m_{H,\alpha,i} \cdot s_{H,\alpha}(\mathbf{x_i}). \tag{3}$$

Once the linear function approximation of the activation function is fully defined, we then plug the linear function coefficients of $s_{H,\alpha}(\mathbf{x})$ inside $q_{H,\alpha,i}(s)$. The result is a linear combination of linear functions, which can also be expressed as a linear function, which we denote by $r_{H,\alpha,i}(\mathbf{x})$,

$$r_{H,\alpha,i}(\mathbf{x}) = q_{H,\alpha,i}\left(s_{H,\alpha}(\mathbf{x})\right), \tag{4}$$

this is a new linear function that relates the output of the neuron $\alpha$ of layer $H$ with the input dimensions around the vicinity of $\mathbf{x_i}$. We then store the coefficients of $r_{H,\alpha,i}(\mathbf{x})$ in a row of the matrix $\mathbf{M_H}$ and move on to the next neuron of layer $H$ repeating the same process.

In the next layer $H - 1$, we will be repeating a similar procedure. We start considering a neuron $\beta$ of this layer, but instead of using $x_j$ as the placeholder argument of $s_{H-1,\beta}(\mathbf{x})$, we plug the corresponding $r_{H,j,i}(\mathbf{x})$, which we can retrieved from the matrix $\mathbf{M_H}$, so that

$$s_{H-1,\beta}(\mathbf{x}) = b_{H-1,\beta} + \sum_{j=1}^{N_H} w_{H-1,\beta,j} \cdot r_{H,j,i}(\mathbf{x}). \tag{5}$$

However, the linear functions $r_{H,j,i}(\mathbf{x})$ contain inside the placeholder variables of $x_j$, which again we should not substitute by the associated values of the base instance. Instead, we simplify the expression $s_{H-1,\beta}(\mathbf{x})$ by multiplying the $w_{H-1,\beta,j}$ weights times each of the coefficients of the $r_{H,j,i}(\mathbf{x})$ functions, which will result in a simple linear function with the $x_j$ variables. Subsequently, we can repeat the exact same procedure that was performed with the activation function of neuron $\alpha$.

Following this same strategy in every neuron of the network and regrouping the parameters we can obtain the linear output function of each neuron, and then storing the associated coefficients in the corresponding $\mathbf{M}$ matrix. Once we arrive at the output layer of the neural network and we have finished filling the entries of $\mathbf{M_1}$, which has a dimension of $N_1 \times N_{in}$, we would have effectively found the $N_1$ linear functions, of all the output dimensions, that approximate the behavior of the network around the base instance chosen.

**Time Complexity**

This algorithm provides a local explanation for a given instance; therefore, it should be executed once for every datapoint studied. Unlike other post-hoc explainability algorithms available, the front-propagation does not require making any perturbations in the input to estimate a linear function approximation of the network. These alternative techniques require many executions of the entire network, which often make them computationally unattractive for real time explainability requirements. On the other hand, the front-propagation requires a single execution, thus its time complexity is the same as a single inference, $O(\sum_{h=1}^{H} N_h)$.

## 4   Results

We developed and packaged `front-prop`, an optimized version of the front-propagation algorithm, which is publicly available on GitHub, [11]. The current version of the code is able to generate real-time linear function explanations of sequential feed forward models developed in Tensorflow and Pytorch. This version can only tackle models that include dense layers, dropout layers, softmax layers, batch normalization layers, and a variety of popular activation functions (ELU, ReLu, SELU, GELU, sigmoid, tanh, swish, softsign, exponential, hard sigmoid, softplus). Further details on how to download and execute `front-prop` along with a model can be found on the descriptions of the code hosted in GitHub.

We considered three different use cases to test the package developed with the front-propagation algorithm:

- Credit granting (classification task): Model trained in Tensorflow, and data obtained from the University of California Irvine Machine Learning Repository.
- Diabetes prediction (classification task): Model trained in Pytorch, and data obtained from the National Institute of Diabetes and Digestive and Kidney Diseases.
- Temperature prediction (regression task): Model trained in Tensorflow, and data obtained from the University of California Irvine Machine Learning Repository.

In Figs. 4-6 we show the results of applying the front-propagation algorithm after having a well-performing trained neural network model in each of the three use cases. For each of the plots shown in the figures, we applied the following process: First, we considered a base instance. Second, we obtained the associated output generated by the neural network. Then, we applied the front-propagation algorithm to this base instance to obtain a linear function approximation of the model. Logically, if we evaluate the input of the base instance using this linear function approximation, the result should match the output obtained with the neural network in the first place. At this point we want to determine whether the linear function approximation obtained is reliable or not. To do so, we want to study several points that are nearby the base instance and evaluate these both with the model and with the linear function. If the results of these two evaluations match, we could then say that the linear function is indeed a good approximation. To generate the points in the surroundings we added random gaussian noise to each of the input dimensions of the base instance. However, we used a limit on the allowed maximum deviation, which we identify in the code with the "proximity threshold" variable (which can range from 0 to 1). The code developed allows the user to decide how many of these exploration points wants to consider for the plots, in all the figures we set this variable to 1,000. To summarize, each point of the plots represents a given input instance, near the base instance, that was evaluated twice: first using the trained neural network, and then using the linear function approximation that was found by the front-propagation algorithm. The horizontal axis represents the output of the trained neural network, and the vertical axis represents the output of the linear function approximation found by `front-prop`.

The points were colored using the Euclidean distance in the normalized input space with respect to the base instance. In each figure we show three rows: the first is associated to a proximity threshold of 0.1, the second with a proximity threshold of 0.5, and the bottom row represents the case where this threshold is 1, the maximum possible value.

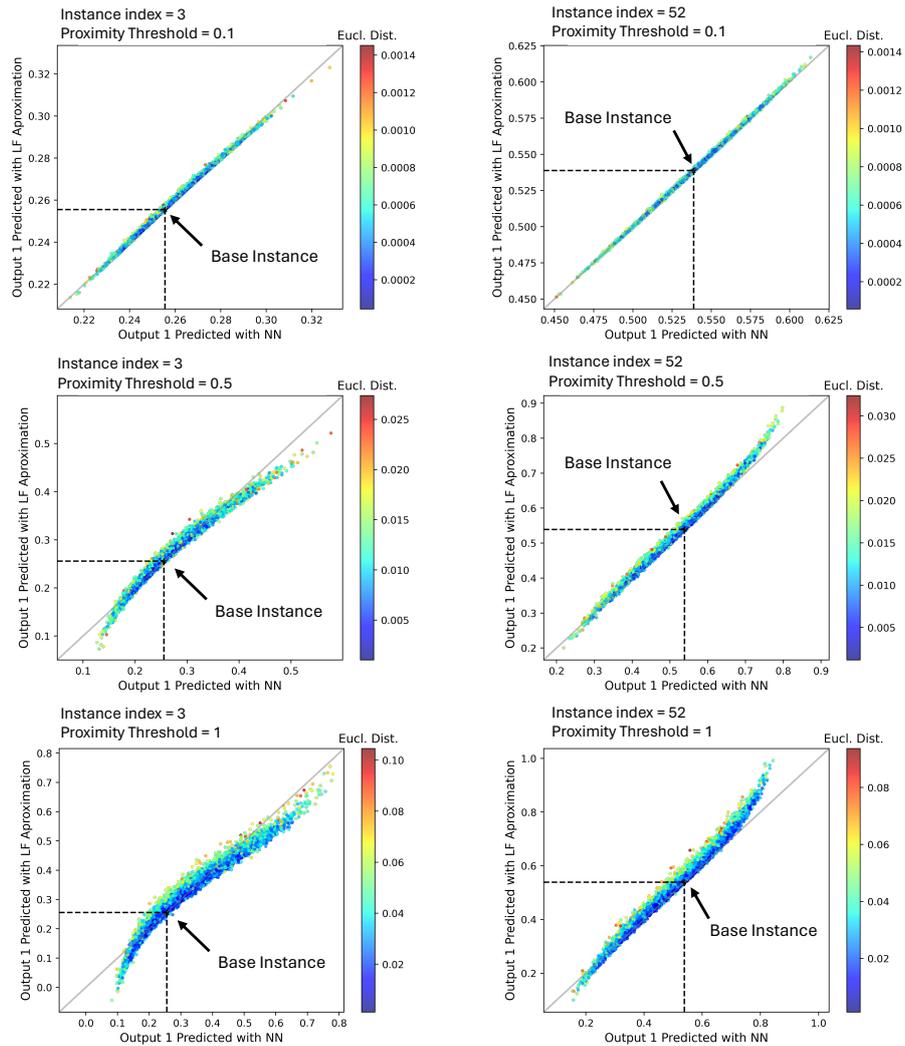

**Fig. 4.** Credit prediction use case: In the top we show a description of the task, the model chosen, and the training hyperparameters. The 6 figures are divided in two columns, each of the columns represent different random base instances: left shows index 3 in the dataset, and right shows index 52 in the dataset.

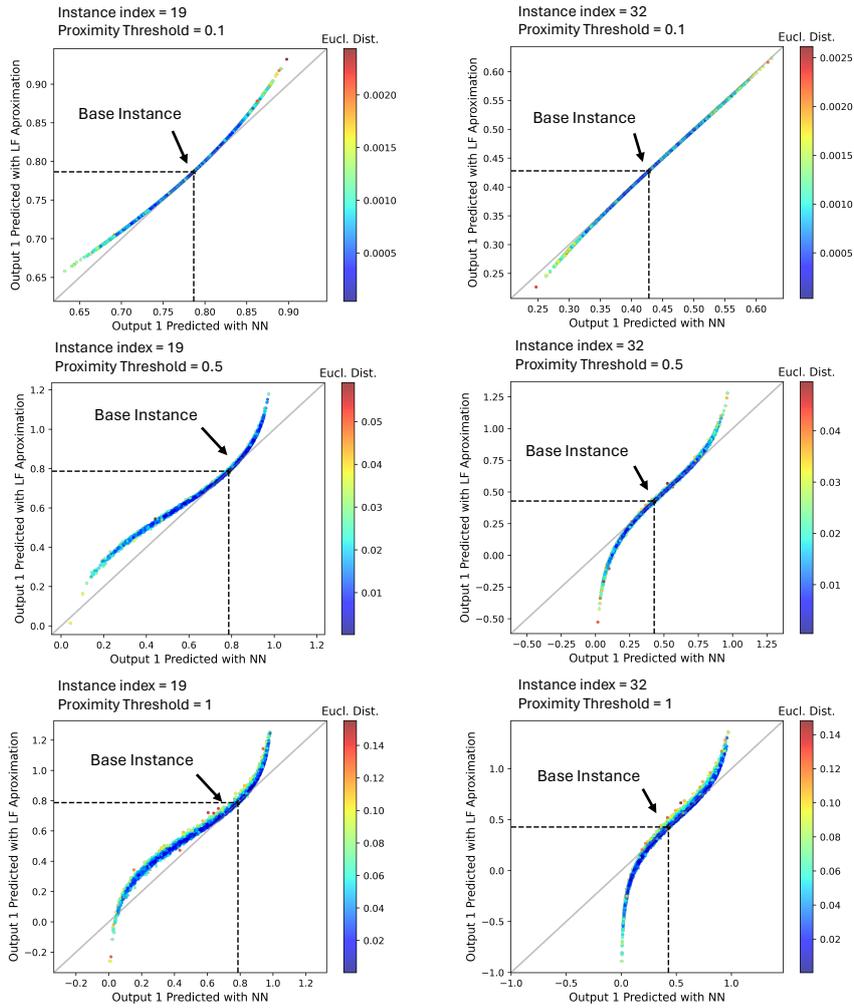

**Fig. 5.** Diabetes prediction use case: In the top we show a description of the task, the model chosen, and the training hyperparameters. The 6 figures are divided in two columns, each of the columns represent different random base instances: left shows index 19 in the dataset, and right shows index 32 in the dataset.

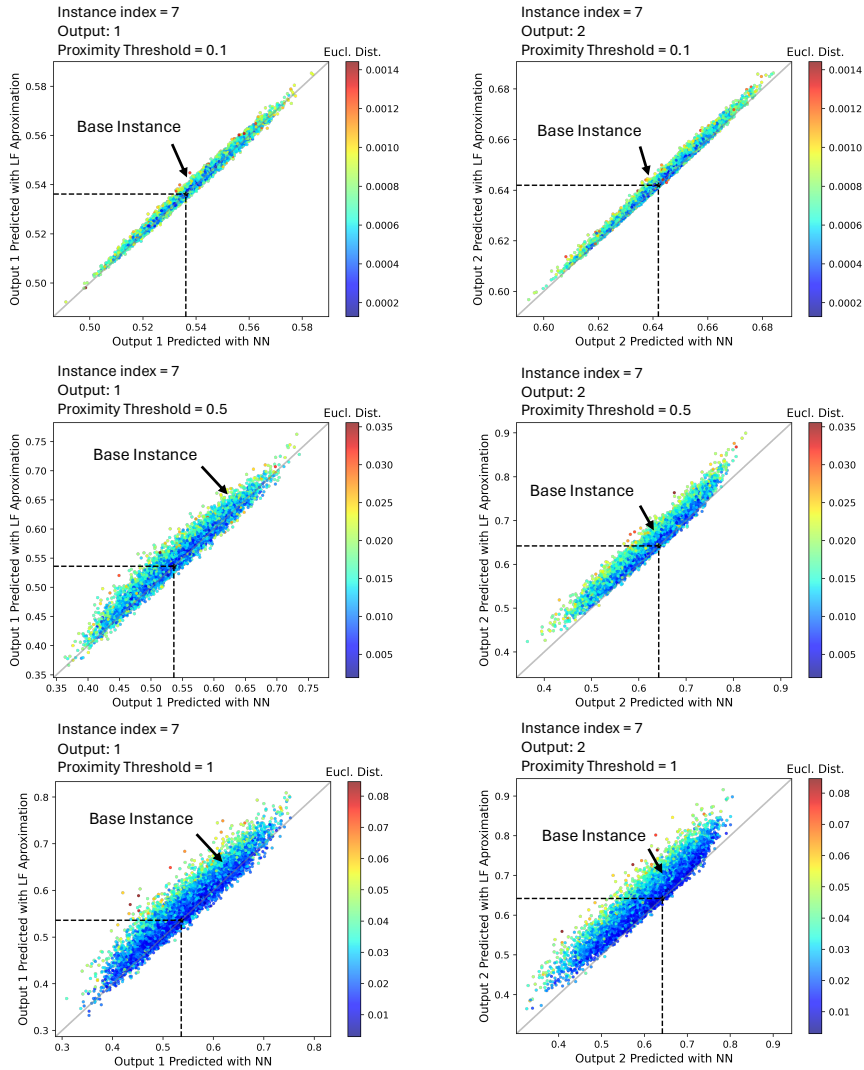

**Fig. 6.** Credit prediction use case: In the top we show a description of the task, the model chosen, and the training hyperparameters. The 6 figures are divided in two columns, each of the columns represent one of the two output dimensions, both columns are showing the same random base instance, which corresponds to index 7 in the dataset.

## 5 Discussion

The scatter of points shown in the plots of Figs. 4-6 exhibit a tangential behavior to the reference diagonal line where the predictions of both the neural network and the linear function match. The tangential pattern is more prominent where the Euclidean distance is smaller. This implies that for instances that around the vicinity of the base instance chosen, the front-propagation algorithm is able to extract a linear function that approximates accurately the behavior of the neural network. This trend was also found in all the other cases we studied. Noticeable, the scatter of points converges to the reference diagonal when we use smaller values of the proximity threshold.

To summarize, the obtention of a linear function approximation of the model is particularly useful for three main reasons:

- If we multiply the input values of the base instance times the corresponding coefficients of the linear function approximation the resulting terms could be seen as the contributions of each input dimension towards the generation of the output. These contributions are particularly interesting to detect potential biases towards certain input dimensions when they exceed a given threshold, or to identify the variables that are mostly influencing the model's outcome when aggregating the contributions across an entire dataset.
- This linear function serves as an explanation of the underlying reasoning of the model. Indeed, in certain use cases one can find what are the "common" reasonings of the network by clustering the coefficients of these linear functions, which then can be associated to certain operational modes of the network. Furthermore, if those "normal" modes are quantified, one could also develop a simple algorithm that classifies when the network's reasoning is becoming too different than those expected, and therefore it could be seen as an outlier reasoning. This is particularly useful in safety critical applications, where the user could decrease the trust in the outputs generated if those have outlier reasonings associated, or in other words if the linear function did not follow one of the common reasoning modes.

Other eXplainable AI algorithms rely on making variations on the input dimensions and subsequently execute the model to estimate the influence of each input dimension in the output. These perturbations-based techniques have a variety of drawbacks:

- The results are different in every execution, because they depend on a random component as a result of the perturbations.
- The linear functions obtained are not the true reasoning followed by the network but an approximation of this.
- The computational cost is very high due to the multiple inferences required and they can easily saturate the resources if the model has many input dimensions, or simply make the problem unsolvable.

On the other hand, the front-propagation algorithm has the following benefits:

- The result is the same no matter the execution because it does not depend on any random component. The solution to this problem is deterministic.
- The linear function obtained is the underlying true reasoning of the network: Not only because this is a valid approximation of the network for points nearby the base instance, but also because the output obtained with this function equals the output of the network when evaluating the base instance (which does not occur in the mentioned perturbation-based methods).
- The computational cost is extremely small compared to the perturbation-based methods because it only requires one forward pass in the network to obtain the result, whereas the aforementioned methods may require thousands, if not more, of inferences.

## 6  Conclusion

We introduced the front-propagation algorithm, which aims to extract the linear function approximations to explain the inner logic of a trained neural network. The main advantage of this algorithm compared to other explainable AI techniques is the requirement to execute just a single forward pass of the network to obtain such linear function. Other methods often require multiple inference executions and do not always provide the same solution. In essence, the front-propagation algorithm provides a significant improvement in computational cost compared to the state of the art, because its time complexity does not depend on multiple runs of the network.

We packaged a Python version of this algorithm in a software called `front-prop`, which is publicly available in GitHub. We also tested `front-prop` in three different uses cases and demonstrated with visual figures the ability to obtain reliable linear function approximations of the models.

We hope this offers researchers and practitioners a fast method to set the pathway for real-time explanations of neural network models.

**Code Availability**

The code of `front-prop` is publicly available in GitHub [11].